\documentclass{article}

\usepackage[final]{neurips_2019}

\usepackage[utf8]{inputenc}
\usepackage[T1]{fontenc}
\usepackage{hyperref}
\usepackage{url}
\usepackage{booktabs}
\usepackage{amsfonts}
\usepackage{nicefrac}
\usepackage{microtype}
\usepackage{graphicx}
\usepackage{xcolor}
\usepackage{lipsum}
\usepackage{amsmath}
\usepackage{graphicx}
\usepackage{hyperref}

\usepackage{float} 

\title{Deep Learning Approaches for Blood Disease Diagnosis Across Hematopoietic Lineages
 \\
  \vspace{1em}
  \small{\normalfont Stanford CS229 Project} 
}

\author{
  Gabriel Bo, Justin Gu, Christopher Sun \\
  Department of Computer Science \\
  Stanford University \\
  \texttt{\{gabebo, justingu, chrisun\}@stanford.edu} 
}

\begin{document}

\maketitle

\begin{abstract} We present a foundation modeling framework that leverages deep learning to uncover latent genetic signatures across the hematopoietic hierarchy. Our approach trains a fully connected autoencoder on multipotent progenitor cells, reducing over 20,000 gene features to a 256-dimensional latent space that captures predictive information for both progenitor and downstream differentiated cells such as monocytes and lymphocytes. We validate the quality of these embeddings by training feed-forward, transformer, and graph convolutional architectures for blood disease diagnosis tasks. We also explore zero-shot prediction using a progenitor disease state classification model to classify downstream cell conditions. Our models achieve greater than 95\% accuracy for multi-class classification, and in the zero-shot setting, we achieve greater than 0.7 F1-score on the binary classification task. Future work should improve embeddings further to increase robustness on lymphocyte classification specifically. \end{abstract}


\section{Introduction and Related Work} \label{intro}
\begin{figure}[H]
    \centering
    \includegraphics[width=0.9\linewidth]{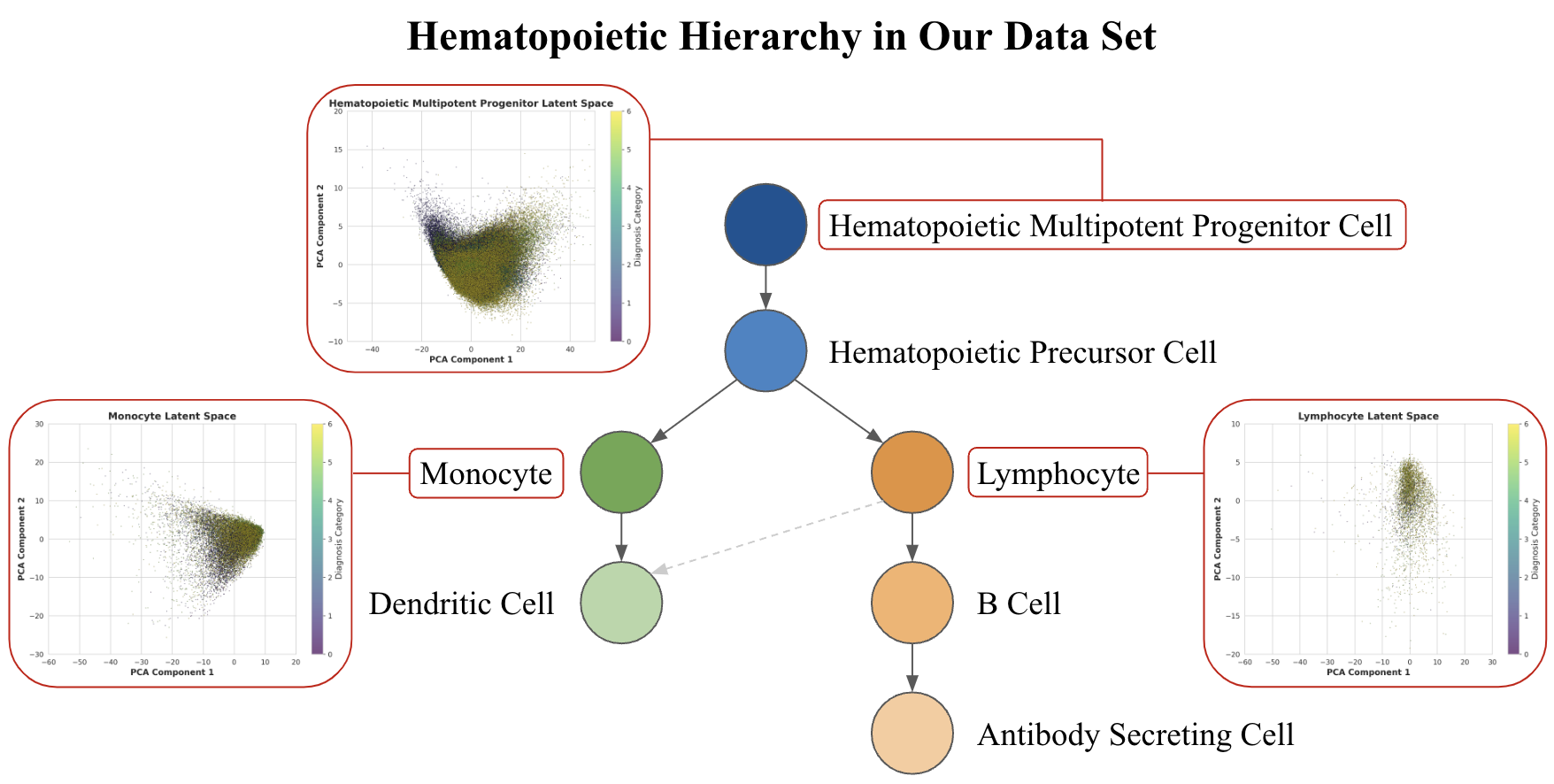}
    \caption{We used single-cell sequencing data from Hematopoietic Multipotent Progenitor Cells, Monocytes, and Lymphocytes in this analysis. The PCA-transformed latent embeddings for these cell types are plotted accordingly.}
    \label{fig:data}
\end{figure}

The differentiation of hematopoietic cells from multipotent progenitors is fundamental to both developmental biology and disease pathology. Recent studies have showcased machine learning’s power to decode complex gene-expression patterns, shedding light on stem cell renewal, aging, and disease progression \citep{arai} \citep{muhsen2020machine}. For example, \citet{wang2023deep} and \citet{cheng2020new} demonstrate the utility of machine learning in this domain. In addition, work by \citet{arai} and related interpretability analyses by \citep{muhsen2020machine} underscore the value of approaches such as self-attention and graph-based methods \citep{graph-nn}, such as a recent use of transformers for detecting skin lesions \citep{attention} \citep{attention-impact}. Such machine learning approaches are useful in their niche area of biology, but do not have the robustness needed to learn generalizable features along a stem cell lineage. While single-cell profiling and deep learning are accelerating biological innovation, the intrinsic heterogeneity within and across cell types continues to be an analysis challenge. Our work is motivated by the need for a foundation modeling framework that is able to learn useful biology along the lineage of hematopoietic stem and progenitor cells (Figure \ref{fig:data}), enabling blood and bone marrow disease diagnosis through purely hematological features.

We utilize high-dimensional gene expression data comprising 29,150 features from hematopoietic cells. An autoencoder compresses this data into a 256-dimensional latent space that contains maximally predictive information generalizable across the hematopoietic lineage \citep{autoencoder}. The autoencoder is trained on upstream multipotent progenitor cells and then applied to downstream cells such as monocytes and lymphocytes. These latent embeddings serve as inputs to various downstream architectures -- including fully connected neural networks, multi-head self-attention models, and graph convolutional networks -- to predict disease states. We further explore a zero-shot classification scenario, where an upstream-trained classifier is evaluated on downstream cell types.

\section{Dataset and Features} \label{data}
Our work leverages the public dataset ``Human circulating hematopoietic stem and progenitor cells in aging, cytopenia, and MDS'' from the Chan Zuckerberg Foundation \citep{dataset}. This dataset comprises over a million cells across nine hematopoietic cell types and provides gene expression measurements for 29,150 genes. Each cell is labeled with one of seven disease states, including 1) myelodysplastic syndrome (MDS), 2) cytopenia, 3) chronic myelomonocytic leukemia (CMML), 4) smoldering multiple myeloma and MDS, 5) myeloproliferative neoplasm and MDS, 6) CMML and MDS, and 7) normal condition. The data set included 199,764 hematopoietic multipotent progenitor cells, 20,643 monocytes, and 4,224 lymphocytes.

\section{Methods} \label{methods}
\subsection{Fully-Connected Autoencoder} \label{approach: autoencoder}
To extract a low-dimensional representation from high-dimensional gene expression profiles, we use a dense autoencoder with an encoder–decoder structure. Let \(\mathbf{x} \in \mathbb{R}^{G}\) be the raw input vector of \(G\) gene expression levels; the autoencoder learns the mapping:
\begin{equation}
    \mathbf{z} = f_{\text{enc}}(\mathbf{x}) \quad \text{and} \quad \hat{\mathbf{x}} = f_{\text{dec}}(\mathbf{z}),
    \label{eq:autoencoder}
\end{equation}
by minimizing reconstruction error, where \(\mathbf{z} \in \mathbb{R}^{d}\) is the latent embedding (with \(d=256\) in our experiments) and \(\hat{\mathbf{x}}\) is the reconstructed input.

The encoder reduces the dimensionality from \(G\) to \(d\) using fully connected layers with Batch Normalization and ReLU activations, enforcing a compact latent space that captures salient features. The decoder mirrors this architecture to reconstruct \(\mathbf{x}\) from \(\mathbf{z}\) with similar layers, culminating in a final output layer. By minimizing the mean squared error between \(\mathbf{x}\) and \(\hat{\mathbf{x}}\), the autoencoder learns a compact representation that is subsequently used to enhance downstream classification tasks across various hematopoietic cell types. 

\subsection{Fully-Connected Neural Network} \label{approach: nn}
To capture non-linear relationships in the latent space and learn a disease diagnosis, we employ a feed-forward network with multiple fully connected layers. Let \(\mathbf{x} \in \mathbb{R}^d\) denote the input vector. The network computes:
\begin{equation}
    \mathbf{h}_1 = \mathrm{ReLU}(\mathbf{W}_1 \mathbf{x} + \mathbf{b}_1), \quad
    \mathbf{h}_2 = \mathrm{ReLU}(\mathbf{W}_2 \mathbf{h}_1 + \mathbf{b}_2), \quad
    \mathbf{y} = \mathbf{W}_3 \mathbf{h}_2 + \mathbf{b}_3.
    \label{equation: nn}
\end{equation}
Key design elements include ReLU activations for non-linearity, batch normalization for training stability, and dropout regularization to mitigate overfitting. The output \(\mathbf{y}\) is subsequently passed through a softmax (multi-class) or sigmoid (binary) layer for classification. This architecture effectively models complex decision boundaries and captures nuanced patterns in gene expression embeddings, outperforming simpler approaches like logistic regression.

\subsection{Transformer Model with Multi-head Self-Attention} \label{approach: transformer}
We further explore advanced architectures using a simplified attention-based model \citep{attention}. Each latent embedding is treated as a sequence of length 256 and dimension of 1. The model begins by projecting the \(d\)-dimensional input to a hidden dimension \(D\) using a linear layer. Next, multi-head self-attention computes queries, keys, and values, capturing diverse feature-to-feature relationships:
\begin{equation}
    \mathrm{Attention}(\mathbf{Q}, \mathbf{K}, \mathbf{V}) = \mathrm{softmax}\Bigl(\frac{\mathbf{Q} \mathbf{K}^\top}{\sqrt{D}}\Bigr)\mathbf{V}.
    \label{equation: attention}
\end{equation}
By splitting the hidden dimension into multiple heads, the model attends to various subspaces:
\begin{equation}
    \mathrm{MultiHead}(\mathbf{Q}, \mathbf{K}, \mathbf{V}) = \bigl[\mathrm{head}_1 \,\|\, \mathrm{head}_2 \,\|\, \dots \,\|\, \mathrm{head}_h\bigr] \, \mathbf{W}^O.
    \label{equation: multihead}
\end{equation}
A small stack of fully connected feed-forward layers then refines the attention outputs, with the final output passed through a classification layer to predict disease labels. This approach effectively captures global, context-aware features within the latent space.

\subsection{Graph Convolutional Networks} \label{approach: graph}
We implement Graph Convolutional Networks (GCNs) \citep{gcn} to assess whether the explicit modeling of cell-to-cell relationships enhances diagnostic ability. The embedded data is represented as a graph, where nodes are samples and edges are formed based on pairwise cosine similarity with thresholding and a fixed maximum to manage memory. Our two-layer GCN updates node representations by aggregating neighbors' features weighted by a normalized adjacency matrix, applying the following propagation rule:
\begin{equation}
    H^{(l+1)} = \sigma \left( \hat{D}^{-\frac{1}{2}} \hat{A} \hat{D}^{-\frac{1}{2}} H^{(l)} W^{(l)} \right),
\end{equation}
where \( H^{(l)} \) is the node feature matrix at layer \(l\), initialized as the input feature matrix \( X \), \( \hat{A} = A + I \) is the adjacency matrix with self-connections, \( \hat{D} \) is the diagonal degree matrix of \( \hat{A} \), \( W^{(l)} \) is the trainable weight matrix at layer \( l \), and \( \sigma \) is an activation function (ReLU in this case).

\section{Experiments and Results}
\subsection{Experiments}
All models were trained using a batch size between 512 and 16,384, depending on memory constraints. We employed a 70/30 train/test random split and used Adam optimization with a learning rate between $3\times10^{-4}$ and $1\times10^{-3}$. Further hyperparameter details are given in the following sections.

\begin{figure}[H]
    \centering
    \includegraphics[width=0.9\linewidth]{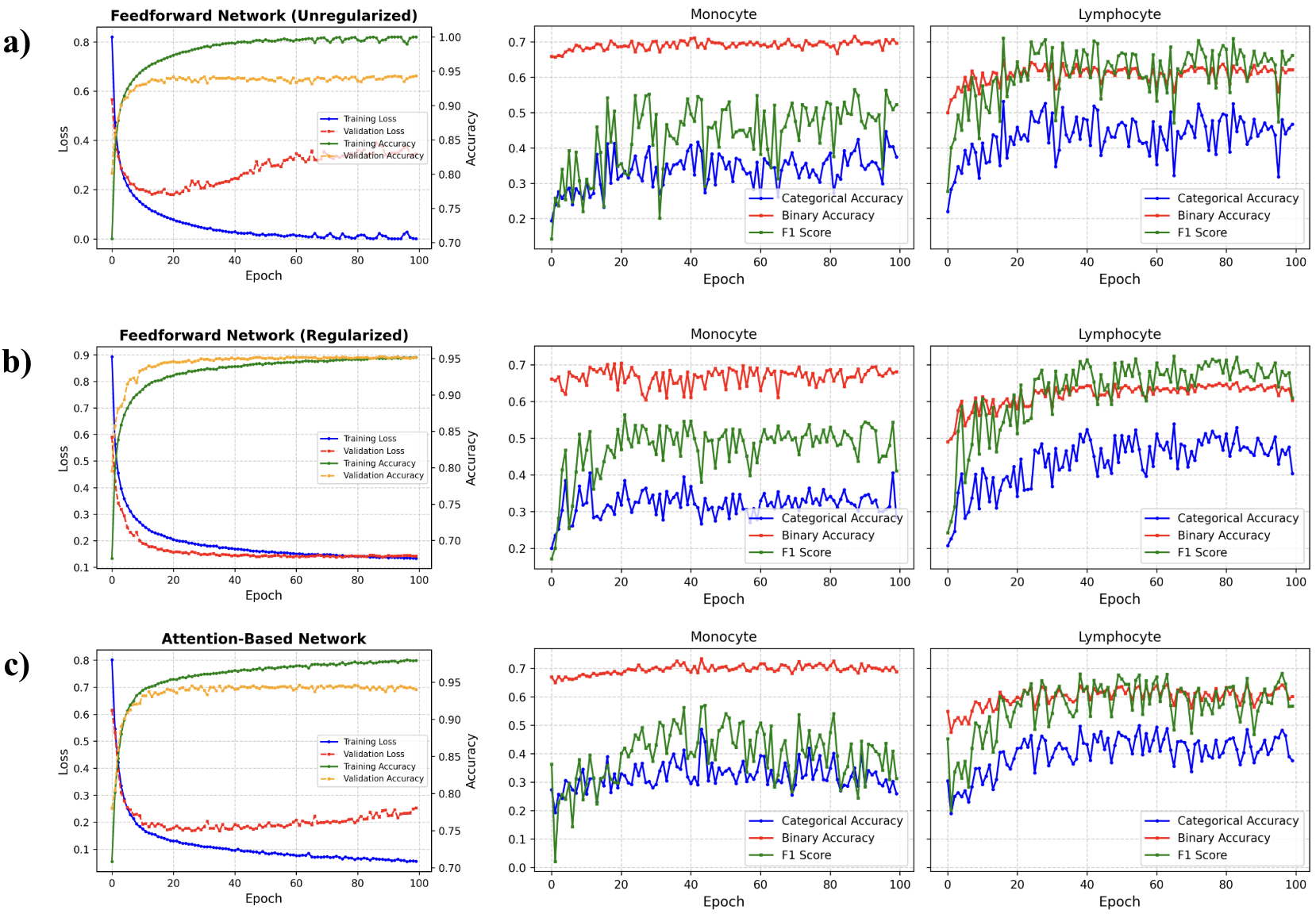}
    \caption{We trained different architectures on hematopoietic multipotent progenitor cells and simultaneously plotted generalization capacity to downstream disease classification in monocytes and lymphocytes.}
    \label{fig:results}
\end{figure}

\paragraph{Fully-Connected Neural Network}
The fully-connected feed forward network, based on the architecture described in Section \ref{approach: nn}, was experimented based on it's dropout rate \citep{ffn}. We first used dropout of 0, which led to some overfitting, but as we increased dropout to 0.1 and 0.2, we noticed that train and test loss grew similar, demonstrating effective mitigation of overfitting, as seen in panel (b) of \ref{fig:results} The model achieving the highest classification accuracy was trained witt 0.1 dropout.

\paragraph{Transformer Model with Multi-head Self-Attention}
We conducted two experiments on the attention-based model: varying the number of attention heads and scaling the hidden dimensions. Using Equation \ref{equation: multihead}, we trained the Transformer classifier with different head counts (e.g., 1, 2, 4, 8) while keeping other hyperparameters fixed. Each head captures a unique projection of the embedding, revealing distinct feature relationships; however, too many heads led to overfitting and longer training times. We found that using 4 heads provided an optimal balance.

In addition, we experimented with the hidden dimension size. While larger dimensions can offer richer representations, they also risk over-parameterization, while smaller dimensions may fail to capture sufficient data variance. Our results indicate that a hidden dimension of 256 -- same as the autoencoder's latent embedding dimension -- effectively prevents overfitting while preserving essential information.

\paragraph{Graph Convolutional Network}
To construct graph representations of the embedded data while staying within memory constraints, we set the cosine similarity threshold to 0.4 and the maximum edges to 1,000. We found that using a hidden dimension size of 128 and a dropout rate of 0.3 yielded the best performance.

\begin{table}[ht]
    \centering
    \caption{Classification Results on Hematopoietic Multipotent Progenitor Cells}
    \begin{tabular}{lccccc}
    \toprule
    \textbf{Model Type} & \textbf{Train Accuracy} & \textbf{Test Accuracy} & \textbf{Monocytes} & \textbf{Lymphocytes}\\
    \midrule
    FFN (dropout 0.0) & 100.00\% & 94.20\% & 0.6945 & 0.6122\\
    FFN (dropout 0.1)  & 95.03\%  & 95.24\% & 0.6912 & 0.6409\\
    Transformer Self-Attention & 97.19\% & 94.59\% & 0.6727 & 0.5126\\
    \bottomrule
    \end{tabular}
    \label{tab:general-classification} 
\end{table}

\paragraph{Improving Downstream Classification Diagnosis} 
Initially, we trained the above architectures on the embeddings for hematopoietic multipotent progenitor cells only. This yielded the learning curves in \ref{fig:results}. Simultaneously, we evaluated the zero-shot performance of the model on downstream monocyte and lymphocytes. The classification of the progenitor cells was robust on its own, as seen in Table \ref{tab:general-classification}, but we also wanted to assess its applicability to downstream cells in a zero-shot setting. The baseline performance yielded binary classification scores of 0.6416 for lymphocytes and 0.7252 for monocytes, and exceeded 0.7 F1-score for lymphocytes. Although these results are not clinical-grade by any means, they demonstrate that progenitor cell genetic profiles can indeed provide a foundation for predicting disease in differentiated cells, and that there is inherent shared information along the hematopoietic lineage that our autoencoder and downstream classifiers learned. 

\begin{table}[ht]
    \centering
    \caption{Classification Results on Downstream Cell Types}
    \begin{tabular}{lcccc}
    \toprule
    \textbf{Model Type} & \multicolumn{2}{c}{\textbf{Monocytes}} & \multicolumn{2}{c}{\textbf{Lymphocytes}} \\
    \cmidrule(lr){2-3} \cmidrule(lr){4-5}
     & \textbf{Train Accuracy} & \textbf{Test Accuracy} & \textbf{Train Accuracy} & \textbf{Test Accuracy} \\
    \midrule
    FFN (dropout 0.1) & 89.84\% & 83.01\% & 74.09\% & 63.41\% \\
    Self-Attention & 90.31\% & 83.05\% & 83.59\% & 64.35\% \\
    GCN & 81.22\% & 79.53\% & 94.11\% & 65.33\% \\
    \bottomrule
    \end{tabular}
    \label{tab:downstream-classification}
\end{table}
Consequently, to improve classification results on monocytes and lymphocytes, we applied the same above architectures (and largely the same hyperparameters) on the embeddings for these cell types, rather than training only on the progenitors. Notably, the self-attention transformer improved monocyte prediction to 83.05\%, while lymphocyte classification reached only 64.35\%, surprisingly below the baseline FFN zero-shot results (Table \ref{tab:downstream-classification}). Moreover, the reduced dataset, stemming from latent space dimensionality reduction and computational constraints with the GCN, highlights the challenges of transferring progenitor insights to lymphocyte disease prediction. \footnote{GCN could not train on the large progenitor cell data, so we only used it for downstream cell disease classification.}

Our results show that while progenitor-derived predictions enhance monocyte disease classification, extending these insights to lymphocytes remains challenging. This highlights the need for targeted methods to capture the complex genetic patterns in lymphocytes and points to promising future research directions for improving cross-cell-type generalization. However, promisingly, our zero-shot approach was able to match, and even surpass, the classification accuracy of the models trained directly on lymphocyte latent embeddings.

\section{Conclusion and Future Work}
This paper presents a machine learning framework that uses a fully connected autoencoder to compress over twenty thousand genetic features into a 256-dimensional latent space, capturing key genetic signatures of hematopoietic cell types. Downstream classifiers -- including a fully-connected neural network, a self-attention based transformer, and a graph convolutional network -- enhance monocyte disease classification while highlighting challenges in lymphocyte prediction. Future work will refine these methods to improve cross-cell-type generalization and disease diagnosis, especially on improving lymphocyte disease classification.

\section{Contributions}
\begin{itemize}
    \item \textbf{Gabriel}: Developed the transformer and feed-forward network code, trained the models, contributed to writing the bulk of the paper, and conducted extensive machine learning research.
    \item \textbf{Justin}: Implemented the graph convolutional network, trained the models, authored large parts of the methods section, and conducted key machine learning research.
    \item \textbf{Christopher}: Developed the foundation model and base code, engineered the autoencoder and data extraction pipelines, provided bioinformatics and machine learning expertise, edited the paper, and made the figures.
\end{itemize}

Code is available on \href{https://github.com/JustinGu32/gene-ml}{GitHub}.
\bibliographystyle{acl_natbib}
\bibliography{references}

\end{document}